# Multi-font Multi-size Kannada Numeral Recognition Based on Structural Features

B.V.Dhandra, R.G.Benne and Mallikarjun Hangarge
P.G .Department of Studies and Research in Computer Science,
Gulbarga University, Gulbarga
INDIA
dhandra_b_v@yahoo.co.in, rgbenne@yahoo.com

**ABSTRACT**

*In this paper a fast and novel method is proposed for multi-font multi-size Kannada numeral recognition which is thinning free and without size normalization approach. The different structural feature are used for numeral recognition namely, directional density of pixels in four directions, water reservoirs, maximum profile distances, and fill hole density are used for the recognition of Kannada numerals. A Euclidian minimum distance criterion is used to find minimum distances and K-nearest neighbor classifier is used to classify the Kannada numerals by varying the size of numeral image from 16 to 50 font sizes for the 20 different font styles from NUDI and BARAHA popular word processing Kannada software. The total 1150 numeral images are tested and the overall accuracy of classification is found to be 100%. The average time taken by this method is 0.1476 seconds.*

**KEY WORDS**

OCR, Kannada Numeral, k-NN, structural feature

## 1. Introduction

Automatic numeral recognition has variety of applications in various fields like reading postal zip codes, number plate of vehicle, reading passport number, employee code, reading bank checks, and form processing. Numeral recognition is an important component of character recognition system due to its vide application. The problem of printed multi-font numeral recognition is difficult task due to the variations in font styles of numerals.

The problem of numeral recognition has been studied for decades and many methods have been proposed such as dynamic programming, hidden Markov modeling, neural network, expert system and combinations of all these techniques [1]. Extensive work has been carried out for recognition of characters/numeral in foreign languages like English, Chinese, Japanese, and Arabic. In the Indian context some major works are reported in Devanagari, Tamil and Bengali numeral recognition [2,3].

The feature extraction plays vital role in recognition system. The various feature extraction methods are reported in the literature, like template matching, projection histogram, zoning, geometric moments, and invariant moments [4] to enable for specific applications. Some methods include fuzzy features [5, 6], invariant moments features [6], template and deformable Templates [11, 12], structural and statistical features [7, 6] extraction. Recent work on printed Indian scripts includes Devanagari, Tamil, Malayalam, Bengali characters [10] and Kannada numerals [3, 6]. Dinesh Acharya U *et al* [8] used the 10-segment string, water reservoir, horizontal and vertical strokes, end point features and k-mean clusters to classify the Kannada isolated numerals. This method requires thinning algorithm. U. Pal *et al* [14] have performed zoning, directional chain code and five-fold cross-validation technique for handwritten numeral recognition. They have considered 100 dimensional feature vectors for recognition. Hence, it suffers from space and time complexity. From the literature it reveals that there are methods which are suffers from larger computation time and is mainly due to preprocessing stage, i.e. size normalization, skeletonning or thinning, the recognition accuracy less. Also they fail to meet the desired accuracy when exposed to the different fonts. Hence, it is necessary to develop a method which is independent of font size, font style and thinning with high accuracy and fast recognition rate. This has motivated us to design a simple, efficient and robust algorithm for Kannada numerals recognition system.

In this paper, the four different categories of structural features proposed by different authors are combined to obtain high degree of accuracy for Kannada numerals recognition. The feature set includes Directional density estimation [9], Water Reservoir principle based features [10], Maximum profile distances and filling holes.

The paper is organized as follows: Section 2 contains the method of sampling and preprocessing. Feature Extraction Method is describes in Section 3. The proposed algorithm is presented in Section 4. The Classifications method is the subject matter of Section 5. The experimental details and results obtained are presented in Section 6. Section 7 contains the conclusion part.

## 2. Kannada numerals and pre-processing

Kannada language is one of the four major south Indian Languages spoken by about 50 million people. The Kannada alphabet consists of 16 vowels and 36 consonants and 10 numerals as given in Table 1.

**Table 1: Kannada Numerals**

| English Numerals | 0 | 1 | 2 | 3 | 4 | 5 | 6 | 7 | 8 | 9 |
|---|---|---|---|---|---|---|---|---|---|---|
| Kannada Numerals | 0 | 1 | 2 | 3 | 4 | 5 | 6 | 7 | 8 | 9 |

Data is collected from Nudi and Baraha software for different font sizes from 16 to 50 with different font styles like, BRH-Kannada, BRH-Amerikannada, BRH-Kailasm, BRH-Vijay, BRH-Kasturi, BRH-Bangaluru, BRH-Sirigannada, KGP_kbd, Nudi B-Akshara, Nudi Akshara, Nudi Akshara-01, Nudi Akshara-02, Nudi Akshara-03, Nudi Akshara-04, Nudi Akshara-05, Nudi Akshara-06, Nudi Akshara-07, Nudi Akshara-08 and Nudi Akshara-09. The printed page containing multiple lines of isolated Kannada numeral is scanned through a flat bed HP scanner at 300 DPI and binarized using global threshold stored in bmp file format. The scanning artefacts are removed by using morphological opening operation. A sample of Kannada numerals and corresponding font styles is presented in Table 2.

**Table 2: A sample data set of printed Kannada numerals**

| Sample Numerals | Font type |
|---|---|
| 0123456789 | BRH-Kannada |
| 0123456789 | BRH-Amerikannada |
| 0123456789 | BRH-Kailasm |
| 0123456789 | BRH-Vijay |
| 0123456789 | BRH-Kasturi |
| 0123456789 | BRH-Bangaluru |
| 0123456789 | BRH-Sirigannada |
| 0123456789 | BRH-Kannada Extra |
| 0123456789 | KGP_kbd |
| ೦೧೨೩೪೫೬೭೮೯ | Nudi Akshara-01 |
| ೦೧೨೩೪೫೬೭೮೯ | Nudi Akshara-02 |
| ೦೧೨೩೪೫೬೭೮೯ | Nudi Akshara-03 |
| ೦೧೨೩೪೫೬೭೮೯ | Nudi Akshara-04 |
| ೦೧೨೩೪೫೬೭೮೯ | Nudi Akshara-05 |
| ೦೧೨೩೪೫೬೭೮೯ | Nudi Akshara-06 |
| ೦೧೨೩೪೫೬೭೮೯ | Nudi Akshara-07 |
| ೦೧೨೩೪೫೬೭೮೯ | Nudi Akshara-08 |
| ೦೧೨೩೪೫೬೭೮೯ | Nudi Akshara-09 |
| ೦೧೨೩೪೫೬೭೮೯ | Nudi B-Akshara |
| ೦೧೨೩೪೫೬೭೮೯ | Nudi Akshara |

## 3. Feature Extraction Method

Feature extraction is an important component of any recognition system and in particular Kannada numerals. In this paper a structural features are used for the recognition of numerals. The directional density estimation (4 features), water Reservoir principle based features (4 features), maximum profile distances (4 features) and fill hole density feature are used for the numeral recognition. All features of test image and training images are normalized in the range of 0 to 1by dividing each feature by the maximum value in that vector.

### 3.1 Directional density estimation

The outer directional density of pixels is counted row/column wise until it touches the outer border of the character in the four directions viz. left, right, top, and bottom direction as shown in Fig.2. It also shows the corresponding directional pixels considered in the count as black band area [9].

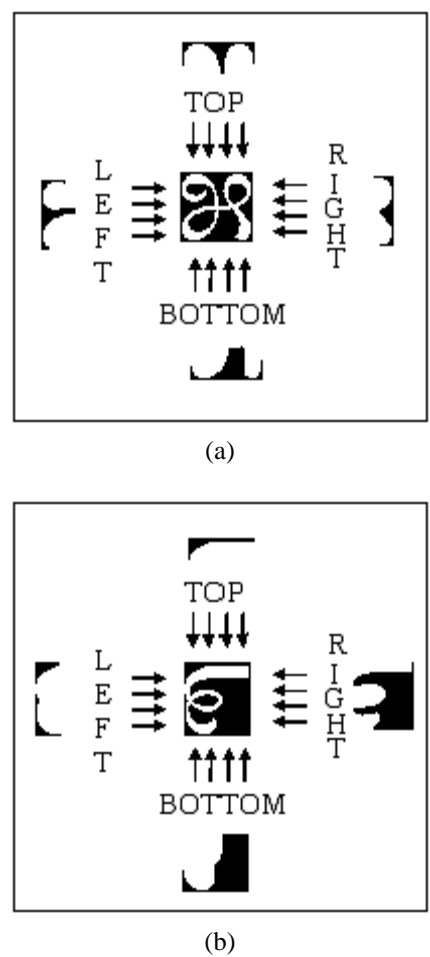

(a)

(b)

**Figure 2: (a)-(b) Directions for Density estimation and pixels consideration**

The ratios of all the directional density of the pixels with the total area are computed and they are stored the feature vector.

### 3.2 Water Reservoir principle based features

The water reservoir principle is as follows. If water is poured from one side of a component, the cavity regions of the component where water will be stored are considered as reservoirs [10].
*Top (bottom) Reservoir*: The reservoir obtained when water is poured from top (bottom) of the component.
*Left (right) Reservoir*: When the water is poured from left (right) side of the component, the cavity regions of the components where water will be stored are considered as left (right) reservoir.

The ratios of the Water Reservoir of the pixels with the total area are computed in four directions and they are stored as the feature vector. Figure 3 illustrates the top, bottom, left, and right reservoir of Kannada numerals.

### 3.3 Fill hole density

The looping area of the numeral is filled with ON pixels [13], further the ratio of fill hole density with total area is estimated and taken as a feature.

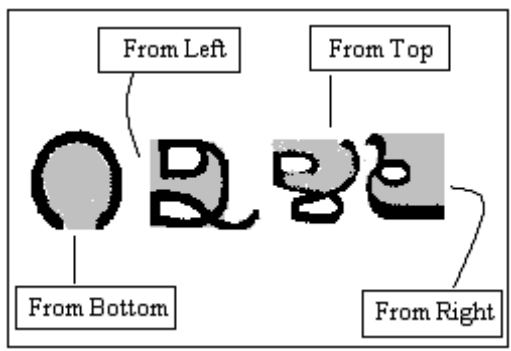

Figure 3: Water Reservoirs in Numerals

### 3.4 Maximum profile distances

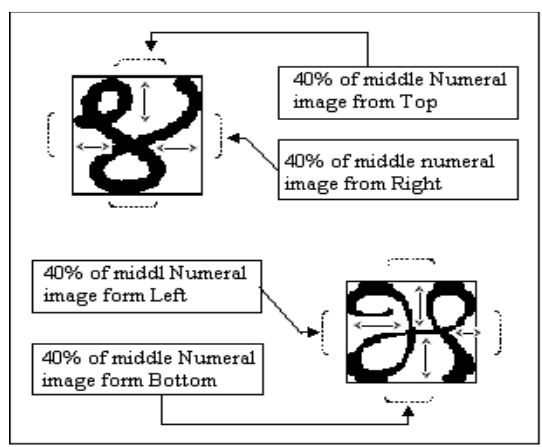

Figure 4: Maximum profile distances from all sides of Bounding Box

After fitting the bounding box on each numeral, their profiles ratio is computed in four directions with the corresponding length. While computing the profile, we have considered only 40% of the middle area in four directions of the bounding box. Thus the ratio of maximum profile is obtained in four directions, the profile feature computations are illustrated in Fig. 4.

### 4. Algorithm

**Input**   : Isolated Binary Kannada Numeral.

**Output**: Recognition of the Numeral.

**Method:** Structural features and k-NN classifier.

1. Preprocess the input image to eliminate the scanning artefacts using morphological opening operation and invert the image.

2. Fit the minimum rectangle bounding box for an input image and crop the digit.

3. Find the directional density of pixels in four direction viz. left, right, top and bottom separately

4. Compute the maximum profile distances from all sides of bounding box and filling hole density.

5. Find the water reservoir in four directions from left, right, top, and bottom.

6. Estimate the minimum distance between feature vector and vector stored in the library by using Euclidian distances.

7. Classify the input image into appropriate class label using minimum distance K-nearest neighbor classifier.

8. Stop.

### 5 Classification

*K-Nearest-Neighbor (KNN) classifier*: Nearest neighbor classifier is an effective technique for classification problems in which the pattern classes exhibit reasonably limited degree of variability. The k-NN classifier is based on the assumption that the classification of an instance is most similar to the classification of other instances that are nearby in the vector space. It works by calculating the distances between one input patterns with  the training patterns. A k-Nearest-Neighbor classifier takes into account only the k nearest prototypes to the input pattern.  Usually, the decision is determined by the majority of class values of the k neighbors.  It however, suffers from the twin problems of high computational cost and memory.
K-nearest neighbor is more general than the nearest neighbor. In other words, nearest neighbor is a special case of k-nearest-neighbor, when k=1. The algorithm is executed for k=1, k=3, k=5 and k=7 the results are compared to find out the optimum value of k. From Table 3 it is clear that k=1 is optimal value.

In the k-Nearest neighbor classification, we compute the distance between features of the test sample and the feature of every training sample. The class of majority among the k-nearest training samples is based on the Euclidian measures.

**Table 3:**
**Error rate using different values of k with KNN classifier by taking different set of training images.**

| NN classifiers with different K values | Number of training samples and their accuracy in percentage (for 1150 test images) | |
|---|---|---|
| | 75 sample | 50 samples |
| K=1 | **100.00** | 99.910 |
| K=3 | 99.912 | 99.389 |
| K=5 | 99.654 | 99.002 |
| K=7 | 99.384 | 98.157 |

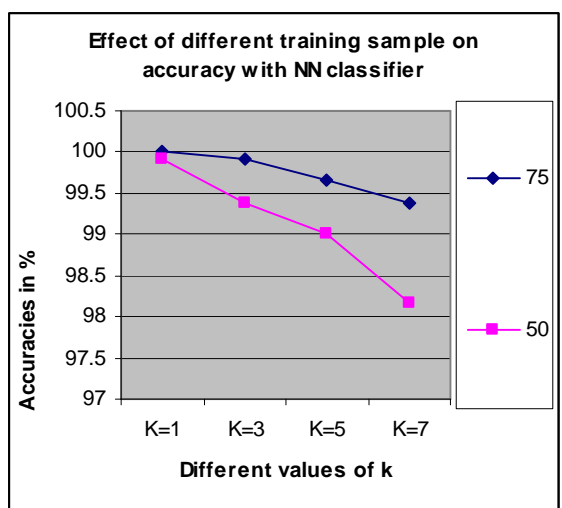

**Figure 5:**
**Effect of different training samples on accuracy by taking different values of k**

**Table 4:**
**Results of numeral recognition for 1150**

| Numeral | Test images | Correctly classified | % Accuracy |
|---|---|---|---|
| 0 | 115 | 115 | 100.00 |
| 1 | 115 | 115 | 100.00 |
| 2 | 115 | 115 | 100.00 |
| 3 | 115 | 115 | 100.00 |
| 4 | 115 | 115 | 100.00 |
| 5 | 115 | 115 | 100.00 |
| 6 | 115 | 115 | 100.00 |
| 7 | 115 | 115 | 100.00 |
| 8 | 115 | 115 | 100.00 |
| 9 | 115 | 115 | 100.00 |
| **Total** | **1150** | **1150** | **100.00** |

## 6 Results and discussion

This algorithm uses 1150 sample of numerals, 20 font styles and 16 to 50 font sizes from Nudi and Baraha Kannada word-processing software. The Table 2 shows the samples of the numerals in different font styled considered under study. For training randomly select 50 images of each numeral with different font styles and sizes are considered. The recognition rate is found to be 100.00% as shown in the Table 4.

## 7. Conclusions

In this paper, thirteen (13) structural features are used for recognition of printed Kannada numerals. In any recognition process, the important problem is to address the feature extraction and suitable classification. The proposed algorithm tries to address both the factors and performs better in terms of accuracy and time complexity. The Overall accuracy of 100.00% is achieved in the recognition process. The proposed method is thinning free, free from size normalization, fast, accurate, and independent of size and font. This work is carried out as an initial attempt, and the aim of the paper is to facilitate for robust Kannada OCR

## References

[1] A.L.Koerich, R. Sabourin, C.Y.Suen, "*Large off-line Handwritten Recognition: A survey*", Pattern Analysis Application 6, 97-121, 2003.

[2] A.F.R. Rahman, R.Rahman, M.C.Fairhurst, "*Recognition of handwritten Bengali Characters: A Novel Multistage Approach*", Pattern Recognition, 35,997-1006, 2002.

[3] R. Chandrashekaran, M.Chandrasekaran, Gift Siromaney, "*Computer Recognition of Tamil, Malayalam and Devanagari characters*", Journal of IETE, Vol.30, No.6, 1984.

[4] Ivind due trier, anil Jain, torfiinn Taxt, "*A feature extraction method for character recognition-A survey* ", pattern Recg, vol 29, No 4, pp-641-662, 1996.

[5] Shamic Surel, P.K.Das, "*Recognition of an Indian Scripts Using Multilayer Perceptrons and fuzzy Features*" Proc. Of 6[th] Int. Conf. on Document Analysis and Recognition (ICDAR), Seattle, pp 1220-1224, 2001.

[6] P.Nagabhushan, S.A.Angadi, B.S.Anami, "*A fuzzy statistical approach of Kannada Vowel Recognition based on Invariant Moments*", *Proc. Of 2[nd] National Conf. on Document Analysis and Recognition* (NCDAR-2003), Mandy, Karnataka, India, pp275-285, 2003.

[7] L.Heutte, T.Paquest, J.V.Moreau, Y.Lecourtier, C.Oliver, "*A structural/ statistical feature based vector*

*for handwritten character recognition*", Pattern Recognition, p.629-641, 1998.

[8] Dinesh Acharya U, N V Subba Reddy and Krishnamoorthi, *"Isolated handwritten Kannada numeral recognition using structural feature and K-means cluster"*, IISN-2007, pp-125-129.

[9] B.V. Dhandra, V.S. Mallimath, Mallikargun Hangargi and Ravindra Hegadi, *"Multi-font Numeral recognition without Thinning based on Directional Density of pixels"*, Proc of first IEEE International conference on Digital Information management" (ICDIM-2006) Bangalore, India, pp.157-160, Dec-2006.

[10] U Pal and P.P.Roy, *"Multi-oriented and curved text lines extraction from Indian documents"*, IEEE Trans on system, Man and Cybernetics-Part B, vol.34, pp.1667-1684, 2004.

[11] J.D. Tubes, A note on binary template matching. *Pattern Recognition*, 22(4):359-365, 1989.

[12] Anil K.Jain, Douglass Zonker, *"Representation and Recognition of handwritten Digits using Deformable Templates"*, IEEE, Pattern analysis and machine intelligence, vol.19, no-12, 1997.

[13] R.C.Gonzal, R.E.Woods, *"Digital Image Processing",* Pearson Education, 2002.

[14] N. Sharma, U. Pal, F. Kimura, "Recognition of Handwritten Kannada Numerals", ICIT, pp. 133-136, 9th International Conference on Information Technology (ICIT'06), 2006.